\documentclass[preprint]{article}

\newif\ifpreprint
\newif\ifanon
\newif\ifappendix

\preprinttrue   
\appendixtrue    

\usepackage{lineno,hyperref}
\modulolinenumbers[5]

\usepackage{t1enc}
\usepackage[utf8]{inputenc}
\usepackage{amsmath,amsfonts}

\usepackage{graphicx}
\usepackage{flushend}
\usepackage{multicol}
\usepackage{multirow}
\usepackage{ragged2e}

\usepackage[english]{babel}

\usepackage{longtable}
\usepackage{subcaption}
\usepackage{array}
\usepackage{cleveref}

\usepackage{orcidlink}

\newcolumntype{C}[1]{>{\Centering\arraybackslash}p{#1}}

\renewcommand{\onecolumn}{}
\renewcommand{\twocolumn}{}

\newcommand{\tabletextsize}{\tiny}

\ifpreprint
\usepackage{draftwatermark}
\SetWatermarkText{\bfseries PREPRINT}
\SetWatermarkScale{3.5}
\SetWatermarkLightness{0.9}
\else
\journal{XXXXXXXXXXXXXXXXXX}
\fi

\title{ERS: a novel comprehensive endoscopy image dataset for machine learning, compliant with the MST 3.0 specification}

\ifanon
\author{------------, ------------}

\else
\author{Jan Cychnerski \orcidlink{0000-0003-2733-4599} \and Tomasz Dziubich \orcidlink{0000-0002-7991-9022} \and Adam Brzeski \orcidlink{0000-0002-1005-4315} }
\date{Department of Computer Architecture \\ Gdańsk University of Technology \\ Narutowicza 11/12, Gdańsk, Poland \\[2ex] September 1, 2021}

\fi

\begin{document}

\maketitle

\begin{abstract}

The article presents a new multi-label comprehensive image dataset from flexible endoscopy, colonoscopy and capsule endoscopy, named ERS. The collection has been labeled according to the full medical specification of 'Minimum Standard Terminology 3.0' (MST 3.0), describing all possible findings in the gastrointestinal tract (104 possible labels), extended with an additional 19 labels useful in common machine learning applications.

The dataset contains around 6000 precisely and 115,000 approximately labeled frames from endoscopy videos, 3600 precise and 22,600 approximate segmentation masks, and 1.23 million unlabeled frames from flexible and capsule endoscopy videos. The labeled data cover almost entirely the MST 3.0 standard. The data came from 1520 videos of 1135 patients.

Additionally, this paper proposes and describes four exemplary experiments in gastrointestinal image classification task performed using the created dataset. The obtained results indicate the high usefulness and flexibility of the dataset in training and testing machine learning algorithms in the field of endoscopic data analysis.

\end{abstract}



\section{Introduction}

The importance and extent of potential benefits from the use of sophisticated information systems in medicine give strong motivation for research on new possibilities of supporting medical examinations. One of the significant directions of research is the development of automatic analysis tools, not only for image, but also offline and on-the-fly analyzing video streams. One of the examples of such fields is an endoscopic examinations of a human gastrointestinal (GI) tract, which is a very important medical procedure in the diagnosis of GI diseases. Diseases of the digestive system comprised of 4\% of the total number of causes of death  in 2016 in the EU \cite{EU}. 
Only colorectal cancer patients comprise of 10.2\% of the total number of new cases in 2018 in the world. In Europe, there were near 500 000 new cases of cancer incidence and mortality exceeded over 250 000 \cite {WHO2018}

By enabling the examination of the interior of the digestive tract, cancer can be detected at an early stage, which in turn gives a good chance of an effective treatment. 

The modern field of endoscopy, which has been developing rapidly in the last two decades, is the Wireless Capsule Endoscopy (WCE). This examination involves the patient swallowing capsule measuring ca. 9x26mm, equipped with a camera and a light source. The recorded images are transmitted wirelessly to a receiver located on the waist around the patient's hips. The result of the examination is a video that is then analyzed by an expert. Unfortunately, the specifics of such films make the analysis very time-consuming - for typical, eight-hour recording it usually takes about 1-2 hours. This is the most difficult task that requires the physician to work in continuous concentration and continuous focus on the image \cite{Hewett2010}. The only indications for the doctor are the information provided by the patient during the medical history. Moreover, during a video colonoscopy,  up to 26\% of polyps can be overlooked, depending on: the endoscopist skills, the time of the exam (part of a day), the quality of colon preparation, visibility (some polyps are behind folds and difficult to detect), the size of the polyps (at early stage, some polyps could be flat and difficult to identify from normal intestinal mucosa) \cite{Ramsoekh2010}.

There is, therefore, a need for a decision support system in the analysis process, which would also allow us to shorten the process and save time for the specialist. The system should conduct a preliminary analysis of the film material examined to indicate fragments of the film depicting medical suspicion of lesions.

One of the common problems in GI endoscopy field is the detection of the occurrence of bleeding and polyps.  Commercial solutions offered by the producers are based on Suspected Blood Indicator determination (SBI) that is related in straightway to a level of red color into image. Several robust techniques have been described in the literature to support these processes. The comprehensive review can be found in \cite{Karargyris2010}, but in spite of passing of time none of the mentioned techniques have been widely commercialized.
In this paper, none of deep learning techniques were presented, which nowadays are a growing trend in general data analysis. Classification accuracy of the developed deep learning approaches is quantitatively compared (or even better) with traditional digital image analysis methods  requiring prior computation of handcrafted features, such as statistical measures using gray level co-occurrence matrix, Gabor filter-bank responses, LBP histograms, gray histograms, HSV histograms and RGB histograms, followed by traditional machine learning methods \cite{szczypinski2017qmazda}. Currently, most solutions utilize other ML techniques (so-called conventional ML) among which the leading ones are SVM, Logit-regression. They are able to solve the task with relatively small training data set. However, the modern trend is deep learning (DL) methods and especially Convolutional Neural Networks (CNN). The achieved accuracy is at the level of ca.92\% and sensitivity - ca.85\%. It shows that DL methods can compete with classical methods \cite{Greenspan2016, Shin2016}. The most relevant studies in deep learning for polyp detection and classification in colonoscopy could be found at \cite{Sing2020}.

A large data training set of valuable data plays a crucial role in DL approach but it is frequently hard to access and most researchers in the medical field are tentative to practice open data science for reasons such as the risk of data misuse by other parties and lack of data-sharing incentives. On the other hand, there is no universal protocol to model, compare, or benchmark the performance of various data analysis strategies \cite{Dinov2016}. Data sets containing medical images are hardly available, making reproducibility and comparison of approaches almost impossible. 

In this paper we present a large data set for creating and benchmarking machine learning algorithms that support medical diagnostics. Our goal was to create the data set in the field of GI endoscopy (colono- as well as gastroscopy), both traditional and WCE. We tried to get all classes encountered in Minimal Standard Terminology 3.0 \cite{MST}, a gold standard platform for the multitude of endoscopic report generators for many years.
To our knowledge it is the first such public set in the world.
It consists of over 115 000 images divided into 99 classes with accordance to MST terminology.
The remainder of this paper is structured as follows:  a review of existing atlases and data sets of images and videos in endoscopy is given in section II. Next, our data set is described in details. In section IV we provide a quantitative assessment of deep learning algorithms for a wide range of diseases and artifacts classification. Finally, section V contains a summary and a conclusion.

\section{Related works}

One of the problems occurring in the process of creating automated classifiers based on deep convolutional neural networks (DCNN) is the acquisition of a training data set that has appropriate size \cite{Litjens2017}. In our paper we propose to come to grips with the problem of automated multi-class classification of GI tract diseases through the application of a DCNN architecture. Hence we need a set of at least thousands of images for a given class/disease. Some of them could be generated through augmentation techniques in a synthetic way. However, the quality of the data set is also essential, and it is crucial that all the images and videos are annotated correctly. Another potential issue in machine learning is over-fitting. A diverse data set is therefore recommended to better enable correct disease classification/detection in new data.

Existing web resources are relatively poor. Mostly, they contain dozen of images, usually with one pathology or photos of healthy organs and act as medical atlases. We can give as examples: El Salvador Gastroenterology Atlas with 5086 endoscopic videos \cite{atlas1} and  Gastrolab - US National Endoscopic Database - 1498 images \cite{atlas3}. It is worth noting that many older atlases have disappeared (e.g. the Atlas of Gastrointestinal Endoscopy - Atlanta South Gastroenterology, US and the Atlas of Gastrointerological Endoscopy Departments of Internal Medicine of Finsterwalde Hospital and Aschersleben Hospital, Germany).

Specialized challenges are another more valuable source of endoscopic images and videos (i.e. MICCAI incl. Giana, Kaggle) which provide benchmark data sets that come with expert ground truth contours and standard evaluation measures to assess automated segmentation performance. 
In MICCAI 2015 challenge which covers a polyp segmentation in both colonoscopy images (SD and HD) and the tasks related to WCE image classification, ASU-Mayo data set was used. It is the first, largest, and a constantly growing set of short and long colonoscopy videos, collected and annotated at the Department of Gastroenterology at Mayo Clinic in Arizona. Currently this data set consists of 38 different, pixel-wise annotated videos (20 in training, 18 in testing set). Each image in this database comes with a ground truth image or a binary mask that indicates the polyp region. 20 short training colonoscopy videos of which 10 videos have a unique polyp inside and the other 10 videos have no polyps, HD (1920x1080) and SD (856x480, 712 x 480)  resolutions. Data set consists of 18996 images and a total of 4278 polyp instances, testing set -  17575 in total and 4313 with polyps \cite{Mayo}.

The ETIS-LARIB (ENSEA/CNRS/University of Cergy-Pontoise) data set contains 196 polyp images with a resolution of 1225x966 pixels in HD \cite{ETIS}. All images were extracted from 34 different colonoscopy videos which contain 44 unique polyps. The total number of polyps is 208. These data sets do not contain any “healthy” (normal) images and are usually used for testing.

The last data set used in the above-mentioned challenges was CVC-Clinic collections (subsets) shared by CVC-Clinic research group. It is being constantly developed; now it has reached the size of 11954 images. The first subset is CVC-ColonDB -- a database of annotated image sequences of colonoscopy video with polyps. It contains 15 short colonoscopy videos, coming from 15 different examinations and provides annotations of the region of interest (ROI)  for 300 images selected from all the videos (resolutions 500x574). These images were selected in order to maximise the visual difference between them. The ROIs define the whole area covering the polyp and they are implemented as binary images, with white masks over a black background. The database consists of 1706 different images, divided into four groups: original files (BMP format), polyp masks, non-informative region masks and contour of the polyp masks. The second one is CVC-ClinicDB2015 contains 612 images with associated polyp and background (here, mucosa and lumen) segmentation masks obtained from 31 polyp videos acquired from 23 patients (384x288) \cite{bernal2012}. On the base of these two data sets, the new one CVC-EndoSceneStill was created, which extends the old annotations to account for lumen, specular highlights as well as a void class for black borders present in each image. In the new annotations, background only contains mucosa (intestinal wall) \cite{Vazquez17}. Additionally, it includes a test set with 44 videos from 36 patients and 912 images.

In \cite{cvc3} we could find a new data set, namely CVC--Clinic2017 (or CVC--12k), consists of 11954 images  from 18 videos with one polyp (in total 10025 images) and test set - 18 videos with 18733 images. Each image that contains a polyp comes with pixel-wise annotations in the CVC--356 and CVC--612 data sets. It has to be noted that the ground truth in the CVC--VideoClinicDB data set represents an approximation  of the polyp in the image using ellipses. 

Kvasir is the largest data set in gastroscopy and consists of images, annotated and verified by medical doctors (experienced endoscopists), including several classes showing anatomical landmarks (Z-line, pylorus, cecum), pathological findings (esophagitis, ulcerative colitis, polyps) or endoscopic procedures in the GI tract; contains 8,000 images divided into 8 classes, 1,000 images for each class. In addition, authors provide several sets of images related to removal of lesions, e.g. `dyed and lifted polyp', `dyed resection margins', etc. The data set consist of the images with different resolution from 720x576 up to 1920x1072 pixels and organized in a way where they are sorted in separate folders named accordingly to the content. Some of the included classes of images have a green picture in picture illustrating the position and configuration of the endoscope inside the bowel, by use of an electromagnetic imaging system (ScopeGuide, Olympus Europe) that may support the interpretation of the image. This type of information may be important for further investigations (thus included), but must be handled with care for the detection of the endoscopic findings \cite{kvasir}. 

The next data set is called KiD and is divided into two sets: KiD1 and KiD2. It is an open academic access database of high quality annotated WCE video and images. (resolution of images 360x360 pixels). The first data set (KiD1) consists of a total of 77 images obtained using MiroCam® (IntroMedic Co, Seoul, Korea) capsule endoscopes. These images illustrate various types of abnormalities, including angioectasias, aphthae, chylous cysts, polypoid lesions, villous oedema, bleeding, lymphangiectasia, ulcers and stenoses. The second one (KiD2) comprise of 2371 WCE images. These images illustrate assorted small bowel findings including polypoid, vascular and, inflammatory lesions. This data set also includes normal images from the esophagus, stomach, small bowel and colon. Beside of this, there are three video data sets but they do not include annotations yet. Abnormalities depicted within this data set include 303 vascular (small bowel angiectasias and blood in the lumen), 44 polypoid (lymphoid nodular hyperplasia, lymphoma, Peutz–Jeghers polyps) and 227 inflammatory (ulcers, aphthae, mucosal breaks with surrounding erythema, cobblestone mucosa, luminal stenoses and/or fibrotic strictures, and mucosal/villous oedema) lesion images and 1778 normal images obtained from the esophagus, the stomach, the small bowel, and the colon \cite{kid}.

All mentioned above data sets are available to the public either after registration or sending an inquiry.

An interesting data set was presented by \cite{wang2017}. They used huge data set consisting of 289 colonoscopy videos (156337 images, 60914 with polyps) form 151 patients with polyp history. Moreover the set contains a division into individual types and characteristics of polyps: carcinomatous, adenomatous, hyperplastic, inflammatory and small, isochromatic, flat. There is no information about type of annotations. Unfortunately, this data set is not available in public.

A summary of mentioned data sets was presented in Tab. \ref{tbl1}. As could be seen that vast majority of data sets are related to colonoscopy and we have deficiency of data in remaining categories of GI diseases and artifacts. 

\newcommand{\longcell}{\pretolerance=10 \tolerance=5 \emergencystretch=0pt }


\onecolumn

\tabletextsize
\setlength\tabcolsep{2pt}

\begin{longtable}{C{11mm}|C{8mm}|C{10mm}|C{10mm}|C{24mm}|C{24mm}|C{24mm}}

\caption{Comparison of considered data sets.}\label{tbl1} \\

Data set name & Moda\-lity& Videos / patients & Total number of images& Classes& Type of annotations& Number of normal images \\
\hline
CVC-ColonDB &
colono\-scopy &
13/
13 &
300 
500×574&
1 class (polyps) 300 images &
\longcell
corresponding ground-truth masks, non-in\-for\-ma\-tive region masks, contour of the polyp masks&
0
\\
\hline
CVC-Clinic2015 (CVC612)&
colono-scopy&
29/
23&
1962 384×288&
1 class  (polyps) 612 images &
testing&
1350 nonpolyp-image\\
\hline
CVC-EndoScene Still&
colono-scopy&
44/
36&
 500×574 or 384×288&
1 class 912&
polyp, lumen, background, specularity, border (void)&
\longcell
combines Colon\-DB with Clinic\-2015 into a new data set with explicit divisions for train, test and validation set\\
\hline
CVC-Clinic2017 &
colono-scopy &
18/
N/D&
11954&
1 class (polyps) 10025
polyp detection&
\longcell
graphical, 
imprecise oval shape covering whole polyp (approximated annotation)&
1929 
(second set - test 18 video with 18733 images)
\\
\hline
KiD&
WCE&
47/
N/D&
2500 
360x360
+ 47 video&
Angiectasia, bleeding, inflammations, polyps&
&

\\
\hline
Kvasir v2&
gastro-scopy&
N/A/ N/A&
8000 from 720x576 up to 1920x1072&
\longcell
8 classes: Z-line, pylorus, cecum anatomical landmarks, pathological findings esopha\-gitis, polyps, ulcerative colitis&
image-wise annotations&
N/A
\\
\hline
ETIS-Larib Polyp DB&
colono-scopy&
34/ N/A&
196 HD 1225×966&
\justifying
1 class (polyps)
44 different polyps with various sizes and  appearances. At least one polyp existed in all 196 images,with the total number of polyps being 208&
graphical pixel-wise annotations&
0 polyp image
\\
\hline
ASU-Mayo &
colono-scopy&
20 SD and HD
(+18 test)/ N/A&
18996 HD and SD &
1 class (polyps)  4278&
graphical pixel-wise annotations&
14718
\\
\hline
GIANA 2017&
WCE&
N/A/ N/A&
1198 576x576&
1 (angiodysplasia)&
graphical, pixel-wise annotations&
600
\\
\hline
Nerthus &
colono-scopy&
21 /N/A &
5525 720x576 &
4  (bowel) &
\longcell
classes showing four-score BBPS-defined bowel preparation quality
the position and configuration of the endoscope inside the bowel. &
1350
\\
\hline

\end{longtable}

\normalsize

\twocolumn

As it was shown in Tab. \ref{tbl1} most of data sets are addressed to one morbid entity or clinical findings, most often related to colonoscopy. The number of patients from whom the study comes is very low (up to several dozen). Images (and annotations) illustrate the same or very similar shoot of the disease, often from adjacent images. Most data sets do not provide pixel-wise annotations and the images are marked only with labels. This results in a rapid decrease in the suitability of them to semantic segmentation issues. Succeeding general problem is that several of the existing data sets are cumbersome to use in terms of permission are restricted. 

 The clean and complete data is one of the crucial parts of a good segmentation/object detection system. This means that spending the time to create a high-quality data set is very important and is directly connected to the quality. These motivations drove us to creating a new large and complete data set called ERS (Endoscopy Recommendation System).

\section{ERS dataset details}

\subsection{Dataset contents}

In the MAYDAY 2012 project \cite{Blokus2012}, carried out at the Gdansk University of Technology and the Medical University of Gdańsk (Clinic of Gastroenterology and Hepatology, GUMed), an attempt to create an ensemble of specialized classifiers of images from endoscopic videos was made. The aim of those classifiers, being a part of the MedEye application \cite{Krawczyk2012}, was a multi-class classification and ROI detection to indicate places on the recording where the potential diseases occurred. This task is high-demand by clinicians during off-line WCE video analysis.

In order to accomplish this task we created a complex data set which consists of more than 6800 annotated images coming from retrospective endoscopic GI examinations. All cases were selected and annotated by physicians from the GUMed. We tried to span numerous set of endoscopic diagnosis, using terminology accordingly with Minimal Standard Terminology (MST 3.0) \cite{MST}. We collected and annotated images of 27 different types of colonoscopic findings (terms defined in MST chapter 5.2) and 54 of upper endoscopy findings (terms defined in MST chapter 5.1). The source videos originated from 1271 patient’s examinations (555 and 712 respectively).

In addition to MST-defined terms, in our data set we also included three other useful in (machine learning applications) categories of terms: healthy GI tract tissues, image quality attributes (such as sharp, blur, motion, stool etc.) and images containing blood. Description of all term categories in the data set is presented in Table \ref{tbl2a}.

\onecolumn

\tabletextsize
\begin{longtable}{c|c|c|c}
\caption{Categories of terms in our data set} \label{tbl2a} \\
Category & Category ID & Description & Number of terms \\ \hline
Gastro & \texttt{g} & All terms from MST 3.0 chapter 5.1 (upper endoscopy) & 70 \\
Colono & \texttt{c} & All terms from MST 3.0 chapter 5.2 (colonoscopy) & 34 \\
Healthy & \texttt{h} & Healthy tissues of upper endoscopy and colonoscopy & 7 \\
Blood & \texttt{b} & Information about blood presence in the endoscopic image & 2 \\
Quality & \texttt{q} & Information about quality and artifacts in the endoscopic image & 10 
\end{longtable}
\normalsize

\twocolumn

Our data set is much more sophisticated than all other data sets presented in Table \ref{tbl1}. It contains annotated data of 123 terms, divided into 5 categories. The overall number of images, annotations and patients is presented in Table \ref{tab:data-summary}. The detailed summary is presented in Table \ref{tbl2}. In `Total' column the number of images of specific term is given, whereas number of primary patient's examinations is shown in `Number of patients'. Annotation masks were done by medical experts in a polygon-shaped forms (representing ROI for selected class; later converted to binary masks). For some annotations, mainly including healthy tissue or image quality, no such masks were used. Every annotation was given a corresponding term. Number of such annotated images is shown in sub-column `Precise' of Table \ref{tbl1}. Note, that the table includes several rows with zeros, because our goal is to demonstrate the degree of compliance of our data set with the full MST specification.

\begin{figure}[ht]
\centering
\includegraphics[width=0.8\textwidth]{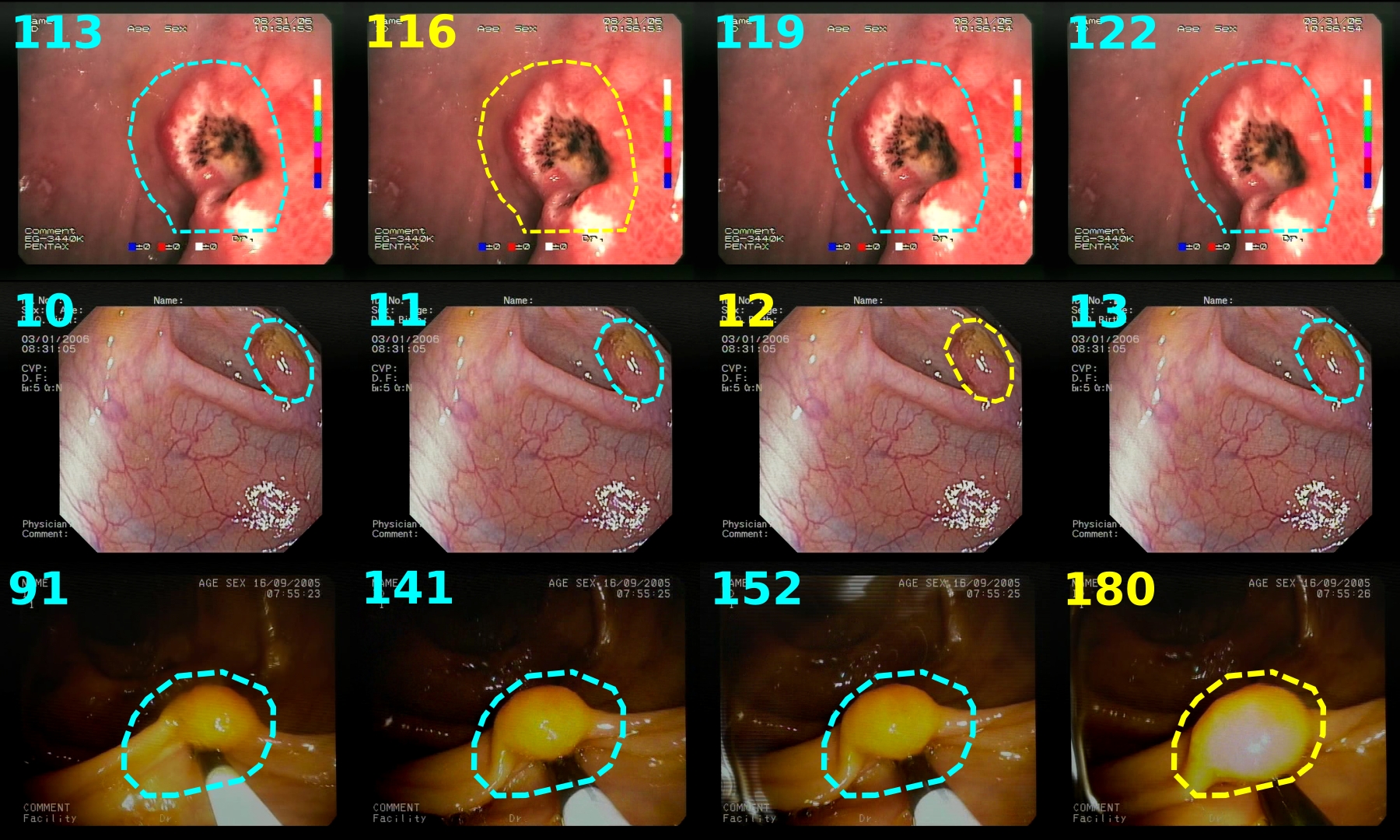}
\caption{Precise (yellow) and imprecise (blue) annotation masks. In corners, frame ID in corresponding video sequence.}
\label{fig:imprecise}
\end{figure}

Due to the fact that most of the images come from videos, which recording speed is about 30 frames per second, in the data set we also included images appearing up to few seconds before or after the annotated source video frame, if they are identical or very similar to the base image, and the annotation mask still fits visually the corresponding region of interest. However, it should be emphasized that this type of annotations/masks was not made by an expert, therefore we have placed them in a separate sub-column (`Imprecise'). Although their lower accuracy, they can be useful in a deep neural network training due to their high abundance. Examples of 'precise' and 'imprecise' annotation masks are presented in \cref{fig:imprecise}. \label{sec:imprecise}

The total number of images in the category may differ from the sum of each term due to the possibility of multiple annotations for one image. Examples of collected images and annotations are presented in Appendixes.

The data set can be obtained at no charge for research purposes from: \newline
\ifanon https://hidden.anonymized.link/ (ANONYMIZED) \else \href{https://cvlab.eti.pg.gda.pl/publications/endoscopy-dataset}{https://cvlab.eti.pg.gda.pl/publications/endoscopy-dataset}\fi

\onecolumn

\tabletextsize

\begin{longtable}{c|c|c|c|c|c|c|c}
\caption{Data set summary: number of images, annotations and patients}\label{tab:data-summary}\\

\multicolumn{2}{c|}{} & \multicolumn{2}{c|}{Images} & \multicolumn{2}{c|}{Annotation masks} & \multicolumn{2}{c}{Number of patients} \\
Modality & Data summary & Precise & Imprecise & Precise & Imprecise & Precise & Imprecise \\
 
 \hline

gastroscopy & Total & 5970 & 982041 & 3606 & 22671 & 1108 & 1119 \\  
\& colonoscopy & With at least one annotation & 5970 & 115429 & 3606 & 22671 & 1098 & 918 \\ 
& With no annotations & --- & 866612 & --- & --- & 10 & 201 \\ 
\hline

WCE & Total & \multicolumn{2}{c|}{366656} & \multicolumn{2}{c|}{---} &  \multicolumn{2}{c}{7}  \\ 

\end{longtable}

\begin{longtable}{c|l|l|c|c|c|c|c|c}
\caption{Terms and number of annotations in our data set}\label{tbl2} \label{tab:terms} \\

\multicolumn{3}{c|}{Terms} &
\multicolumn{2}{c|}{Annotated images} & 
\multicolumn{2}{c|}{Annotation masks} & 
\multicolumn{2}{c}{Number of patients} \\
\hline
Id & Category & Term name & Precise & Imprecise & Precise & Imprecise & Precise & Imprecise \\

\hline

g & gastro & \textit{[total]} & 1779 & 21680 & 1790 & 8035 & 591 & 473 \\ 
g02 &  & achalasia & 9 & 25 & 9 & 25 & 3 & 3 \\ 
g03 &  & barretts esophagus & 20 & 119 & 20 & 119 & 8 & 7 \\ 
g04 &  & benign stricture & 4 & 25 & 4 & 25 & 1 & 1 \\ 
g05 &  & bleeding of unknown origin & 8 & 27 & 8 & 27 & 5 & 4 \\ 
g06 &  & coeliac disease & 7 & 26 & 7 & 26 & 2 & 2 \\ 
g07 &  & crohns disease & 6 & 37 & 6 & 37 & 1 & 1 \\ 
g08 &  & dieulafoy lesion & 9 & 11 & 9 & 11 & 4 & 3 \\ 
g10 &  & duodenal bulb deformity & 5 & 15 & 5 & 15 & 3 & 3 \\ 
g11 &  & duodenal cancer & 19 & 37 & 19 & 37 & 5 & 3 \\ 
g14 &  & duodenal polyp & 27 & 99 & 27 & 99 & 10 & 8 \\ 
g15 &  & duodenal postoperative appearance & 1 & -- & 1 & -- & 1 & -- \\ 
g18 &  & duodenal ulcer & 106 & 436 & 106 & 436 & 44 & 34 \\ 
g19 &  & duodenal ulcer with bleeding & 26 & 87 & 26 & 87 & 12 & 10 \\ 
g20 &  & duodenopathy:erosive & 21 & 146 & 21 & 146 & 8 & 8 \\ 
g22 &  & duodenopathy:hyperemic & 5 & 33 & 5 & 33 & 4 & 3 \\ 
g25 &  & esophageal caustic injury & 5 & 12 & 5 & 12 & 2 & 2 \\ 
g26 &  & esophageal cancer & 54 & 237 & 54 & 237 & 20 & 16 \\ 
g27 &  & esophageal candidiasis & 12 & 144 & 12 & 144 & 3 & 3 \\ 
g28 &  & esophageal diverticulum & 13 & 45 & 13 & 45 & 6 & 4 \\ 
g29 &  & esophageal fistula & 17 & 76 & 17 & 76 & 8 & 8 \\ 
g30 &  & esophageal foreign body & 5 & 1 & 5 & 1 & 2 & 1 \\ 
g31 &  & esophageal polyp & 23 & 99 & 23 & 99 & 11 & 10 \\ 
g32 &  & esophageal postoperative apperance & 1 & -- & 1 & -- & 1 & -- \\ 
g33 &  & esophageal stricture & 23 & 127 & 23 & 127 & 9 & 8 \\ 
g35 &  & esophageal submucosal tumor & 3 & 6 & 3 & 6 & 1 & 1 \\ 
g36 &  & esophageal varices & 170 & 466 & 170 & 466 & 66 & 55 \\ 
g37 &  & extrinsic compression & 7 & 32 & 7 & 32 & 1 & 1 \\ 
g39 &  & gastric cancer & 87 & 274 & 87 & 274 & 34 & 28 \\ 
g40 &  & gastric diverticulum & 1 & 2 & 1 & 2 & 1 & 1 \\ 
g41 &  & gastric fistula & 3 & 7 & 3 & 7 & 1 & 1 \\ 
g42 &  & gastric foreign body & 19 & 65 & 19 & 65 & 10 & 8 \\ 
g43 &  & gastric caustic injury & 2 & 1 & 2 & 1 & 2 & 1 \\ 
g44 &  & gastric lymphoma & 20 & 31 & 20 & 31 & 7 & 5 \\ 
g45 &  & gastric polyp(s) & 192 & 1210 & 192 & 1210 & 72 & 54 \\ 
g46 &  & gastric postoperative appearance & 22 & 112 & 22 & 112 & 11 & 7 \\ 
g47 &  & gastric retention & 18 & 27 & 18 & 27 & 7 & 2 \\ 
g50 &  & gastric ulcer & 204 & 846 & 204 & 846 & 88 & 64 \\ 
g51 &  & gastric ulcer with bleeding & 20 & 94 & 20 & 94 & 7 & 5 \\ 
g52 &  & gastric ulcer:anastomotic & 17 & 24 & 17 & 24 & 10 & 7 \\ 
g53 &  & gastric varices & 82 & 476 & 82 & 476 & 24 & 19 \\ 
g54 &  & gastropathy:erosive & 83 & 446 & 83 & 446 & 31 & 23 \\ 
g55 &  & gastropathy:hemorrhagic & 13 & 46 & 13 & 46 & 3 & 3 \\ 
g56 &  & gastropathy:hyperemic & 33 & 510 & 33 & 510 & 17 & 12 \\ 
g57 &  & gastropathy:hypertrophic & 2 & 12 & 2 & 12 & 1 & 1 \\ 
g59 &  & gastropathy:portal hypertensive & 115 & 14385 & 115 & 631 & 46 & 47 \\ 
g61 &  & hiatus hernia & 13 & 45 & 13 & 45 & 7 & 7 \\ 
g62 &  & mallory:weiss tear & 16 & 43 & 16 & 43 & 8 & 6 \\ 
g63 &  & other esophagitis & 95 & 335 & 95 & 335 & 37 & 31 \\ 
g65 &  & post sclerotherapy appearance & 19 & 71 & 19 & 71 & 9 & 7 \\ 
g66 &  & pyloric stenosis & 22 & 35 & 22 & 35 & 7 & 5 \\ 
g67 &  & reflux esophagitis & 37 & 100 & 37 & 100 & 17 & 8 \\ 
g68 &  & schatzki ring & 4 & 32 & 4 & 32 & 1 & 1 \\ 
g69 &  & scar & 21 & 61 & 21 & 61 & 11 & 6 \\ 
g70 &  & submucosal tumor & 24 & 131 & 24 & 131 & 6 & 4 \\ 

\hline
c & colono & \textit{[total]} & 2199 & 37391 & 1635 & 14636 & 482 & 387 \\ 
c01 &  & angiodysplasia & 61 & 597 & 61 & 597 & 14 & 13 \\ 
c02 &  & bleeding of unknown origin & 5 & 5 & 5 & 5 & 3 & 2 \\ 
c03 &  & colitis:ischemic & 20 & 54 & 20 & 54 & 5 & 4 \\ 
c05 &  & colorectal cancer & 528 & 25345 & 270 & 2333 & 92 & 82 \\ 
c08 &  & crohns disease:active & 127 & 1306 & 127 & 1306 & 19 & 16 \\ 
c10 &  & crohns disease:quiescent & 18 & 151 & 18 & 151 & 11 & 9 \\ 
c11 &  & diverticulitis & 1 & 6 & 1 & 6 & 1 & 1 \\ 
c12 &  & diverticulosis & 83 & 325 & 83 & 325 & 29 & 16 \\ 
c13 &  & fistula & 18 & 217 & 18 & 217 & 5 & 5 \\ 
c14 &  & foreign body & 1 & 4 & 1 & 4 & 1 & 1 \\ 
c15 &  & hemorrhoids & 11 & 65 & 11 & 65 & 5 & 4 \\ 
c16 &  & ileitis & 3 & 13 & 3 & 13 & 1 & 1 \\ 
c17 &  & lipoma & 12 & 113 & 12 & 113 & 3 & 3 \\ 
c19 &  & melanosis & 19 & 75 & 19 & 75 & 6 & 6 \\ 
c20 &  & parasites & 22 & 179 & 22 & 179 & 2 & 2 \\ 
c22 &  & polyp & 950 & 5395 & 629 & 5395 & 247 & 185 \\ 
c23 &  & polyposis syndrome & 16 & 220 & 16 & 220 & 7 & 4 \\ 
c24 &  & postoperative appearance & 10 & 18 & 10 & 18 & 6 & 5 \\ 
c25 &  & proctitis & 9 & 173 & 9 & 173 & 3 & 3 \\ 
c26 &  & rectal ulcer & 22 & 510 & 22 & 510 & 8 & 7 \\ 
c27 &  & solitary ulcer & 12 & 211 & 12 & 211 & 7 & 5 \\ 
c28 &  & stricture:inflammatory & 1 & 6 & 1 & 6 & 1 & 1 \\ 
c29 &  & stricture:malignant & 15 & 131 & 15 & 131 & 9 & 6 \\ 
c30 &  & stricture:postoperative & 2 & 2 & 2 & 2 & 1 & 1 \\ 
c31 &  & submucosal tumor & 2 & 8 & 2 & 8 & 1 & 1 \\ 
c32 &  & ulcerative colitis:active & 162 & 1746 & 162 & 1746 & 33 & 28 \\ 
c34 &  & ulcerative colitis:quiescent & 84 & 773 & 84 & 773 & 35 & 28 \\ 

\hline
h & healthy & \textit{[total]} & 1019 & 19464 & -- & -- & 67 & 110 \\ 
h01 &  & esophagus & 33 & 1006 & -- & -- & 6 & 13 \\ 
h02 &  & stomach & 42 & 1214 & -- & -- & 6 & 29 \\ 
h03 &  & duodenum & 29 & 1154 & -- & -- & 6 & 17 \\ 
h05 &  & small-bowel & 4 & 65 & -- & -- & 1 & 1 \\ 
h06 &  & upper & 9 & 8268 & -- & -- & 2 & 56 \\ 
h07 &  & colon & 902 & 11129 & -- & -- & 49 & 53 \\ 

\hline
b & blood & \textit{[total]} & 814 & 174 & 181 & -- & 133 & 41 \\ 
b01 &  & blood & 184 & 173 & 181 & -- & 77 & 40 \\ 
b02 &  & no blood & 630 & 1 & -- & -- & 56 & 1 \\ 

\hline
q & quality & \textit{[total]} & 974 & 94470 & -- & -- & 97 & 224 \\ 
q01 &  & sharp & 259 & 57530 & -- & -- & 3 & 172 \\ 
q02 &  & blur & 428 & 36922 & -- & -- & 25 & 174 \\ 
q03 &  & bile & 18 & 14 & -- & -- & 6 & 5 \\ 
q04 &  & food & 1 & -- & -- & -- & 1 & -- \\ 
q05 &  & tooclose & 82 & 50 & -- & -- & 23 & 15 \\ 
q06 &  & air & 86 & 48 & -- & -- & 33 & 19 \\ 
q07 &  & defocus & 51 & 25 & -- & -- & 27 & 18 \\ 
q08 &  & light & 60 & 36 & -- & -- & 22 & 17 \\ 
q09 &  & motion & 87 & 50 & -- & -- & 40 & 27 \\ 
q10 &  & stool & 22 & 30 & -- & -- & 10 & 3 \\

\end{longtable}
\normalsize
\twocolumn

\subsection{Dataset structure}

The collection is stored in a form that makes it as easy and versatile as possible to use in a variety of machine learning problems. All visual data are stored as lossless compressed PNG files: endoscopic images as RGB images, annotation masks as monochrome PNG bitmaps. The labels (annotations) corresponding to the frames are included in the file names, and for convenience in the file \texttt{labels.csv}.Precise and imprecise masks and annotations are separated in distinct directories.  Shortened IDs of all labels, along with descriptions are included in the file \texttt{names.json} and \texttt{names.csv}.

\textbf{Patients.} Due to the versatility of the dataset and the large number of unbalanced classes, a priori division of the data into training, validation and test sets is not possible. Such a division must be made by the researcher himself, taking into account the specifics of the problem he is investigating. In order to facilitate the construction of these sets, data are divided according to the patients they originate from. The data from each patient is stored in separate directories -- to avoid bias, we recommend that data from a single patient never be simultaneously included in the training and test sets. The patients are numbered $0000 - 1135$ , and the directories containing the data are named as such.

\textbf{Single frames and video sequences.} The collection contains video sequences from endoscopic examinations and single frames cut from them. For each patient, there is one directory containing all \textit{single} frames (\texttt{samples}) and several directories containing the video sequences (\texttt{seq\_01}, \texttt{seq\_02}, etc.). \textit{Single} frames (samples) are mostly cut from video sequences -- their annotations are \textit{precise} (as defined in \cref{sec:imprecise}). The frames in the movie sequences contain mostly \textit{imprecise} annotations.

\textbf{Endoscopic images.} Each directory \texttt{samples} and \texttt{seq\_N} contains a subdirectory \texttt{frames}, which includes the original endoscopic frames: color RGB images in PNG format. If present, personal information has been anonymized with black rectangles. The filename of each file corresponds to the frame number in the video sequence (e.g., \texttt{seq\_01/frames/000215.png}) or the ordinal number of the cut \textit{single} frame (e.g., \texttt{samples/frames/000001.png}).

\textbf{Annotation files, labels, masks}. If there are annotations for a given patient, the \texttt{samples} and \texttt{seq\_N} directories contain the \texttt{labels} subdirectory. This directory contains all annotations, labels, and masks for a given sequence or single images (samples). If any annotations exist for an image in the \texttt{frames} directory, they are located in the \texttt{labels} directory. Multiple annotation files may exist for each endoscopic image. Each annotation file is named according to convention: \texttt{[frame number]\_[label 1]\_[label 2]\_[label...].png}. The frame number denotes the number of the corresponding annotated endoscopic image from the \texttt{frames} directory. An annotation file may contain multiple different label IDs in the name, separated by an underscore character -- the label IDs are consistent with \cref{tab:terms} and the \texttt{names.csv} file. If the annotation is a mask, it is stored as a monochrome PNG file. If the annotation contains only labels (no mask) it is saved as an empty file (0 bytes in size). As mentioned earlier, the masks in the \text{samples} directories are \textit{precise}, and in the \texttt{seq\_N} directories they are \textit{imprecise}.

\begin{figure}[ht]
\centering
\includegraphics[height=6cm]{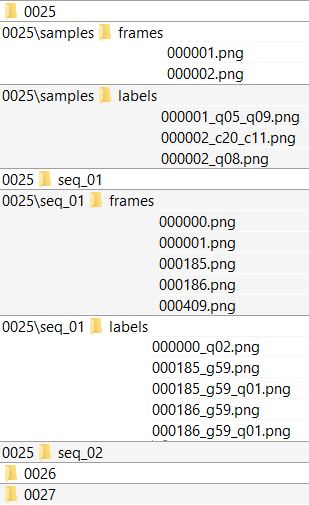}
\caption{Fragment of the dataset directory tree}
\label{fig:tree}
\end{figure}

An excerpt of the dataset directory tree is shown in \cref{fig:tree}. Some exemplary files from the dataset along with an explanation are described below:

\begin{itemize}
    \item \texttt{0025/samples/frames/000001.png} --- cut single frame -- endoscopic image of patient number 25
    \item \texttt{0025/samples/labels/000001\_q05\_q09.png} --- annotation file for the above endoscopic image. It specifies two labels (q05 and q09). The file is empty (size of 0 bytes), so it does not contain an annotation mask.
    \item \texttt{0025/samples/labels/000003\_g59.png} --- annotation file for another endoscopic image {samples/000003.png}. It specifies the g59 label. The file is a monochrome PNG file, so it is an annotation mask of the g59 label. It is contained in the \texttt{samples} directory, so it is a \textit{precise} mask.
    \item \texttt{0025/seq\_01/frames/000185.png} --- endoscopic frame 185 from video sequence 01 for patient 0025. 
    \item \texttt{0025/seq\_01/labels/000185\_g59.png} --- annotation for the above endoscopic frame -- label g59. It is a monochrome PNG file, so it is an annotation mask. It is contained in the \texttt{seq\_N} directory, so it is a \textit{imprecise} mask.
\end{itemize}

\section{Exemplary experiments using the dataset}
\label{sec:experiments}

In order to confirm the usefulness of the described data set, research on the efficiency of classifiers has been carried out. Our algorithms had been implemented in the form of single, universal models of a deep neural network (end-to-end approach). Three popular deep neural network architectures were used: Inception v3 \cite{Szegedy_2016_CVPR},  NASNet-Mobile \cite{Zoph2017} and MobileNet v1 \cite{howard2017mobilenets}. The models were trained to recognize multiple classes at a time (multi-label classification, with threshold 0.5 on the output layer).

Four main problems were taken into account, as presented in Tab. \ref{tbl3}. They reflect the different possible uses of target classifiers. 
The first of them (\textit{10-CLASS}) can be used in screening, the next two in more accurate examination (\textit{5-CLASS}, \textit{2-CLASS}), directed at a specific group of diseases or a binary classification.
The last (\textit{BLUR}) one will allow us to discard blurry images from the recording,  thus reducing the number of images to be viewed.
Each class in the problems was built by combining several relevant terms from the data set (e.g. ‘normal tissue’ class was constructed using all ‘healthy’ terms). Exact recipes  used to create the specific class, and resulting data set sizes were placed in Appendixes. On top of that, two variants of data were used to train the models: (1) only ‘precise’ images, and (2) both ‘precise’ and ‘imprecise’ images altogether. For testing, only ‘precise’ images were used.

\onecolumn

\tabletextsize

\begin{longtable}{C{15mm}|C{42mm}|c|c|c|c|c}
\caption{Description of four considered classification problems and data sets used during training process}\label{tbl3}\\
\multirow{2}{*}{Problem} & \multirow{2}{4cm}{Predicted classes}&\multirow{2}{*}{Variant} &
\multicolumn{2}{c|}{Training set} & 
\multicolumn{2}{c}{Testing set} \\
\cline{4-7}
&&& Images & Patients& Images&Patients\\
\hline
\multicolumn{1}{p{1mm}|}{\vspace{1mm}\multirow{2}{*}{10-CLASS}}&
\multirow{2}{4cm}{normal, blur, polyp, cancer,ulcer, blood, inflammation,\newline crohn, diverticulum, varicose} &
Precise& 4308 & 803 & 782 & 199 \\
\cline{3-7}
\multicolumn{1}{p{1mm}|}{\vspace{2mm}} &&Imprecise & 88779 & 817 & 782 & 199\\
\hline
\multirow{2}{*}{5-CLASS}& 
\multirow{2}{4cm}{normal, polyp, cancer, ulcer, blood} &
Precise & 2954 & 586 & 559 & 133\\
\cline{3-7}
&&Imprecise & 46835 & 610& 559& 133\\

\hline
\multirow{2}{*}{2-CLASS}&
\multirow{2}{*}{normal, disease}&
Precise& 4098& 874& 901& 216\\
\cline{3-7}
&&Imprecise& 63102& 885& 901& 216\\
\hline

\multirow{2}{*}{BLUR} &
\multirow{2}{*}{blur, sharp}&
Precise & 4521& 878& 933& 21\\
\cline{3-7}
&&Imprecise& 90905& 890& 933& 218\\

\end{longtable}

\normalsize

\twocolumn

The data was split so as 80\% of patients belong to ‘train’ subset, and 20\% of patients to ‘test’ subset. 
In each data set, number of images for each class was balanced using repetition and augmentation (rotation, flip, color change, brightness change, blur). The number of patients and images per patient were not balanced, except for an additional 'MobileNet-Balanced' test in which the number of images per patient in each class was proportional to the square root of original number.

\pagebreak  

\begin{figure}[ht]
\includegraphics[width=\textwidth]{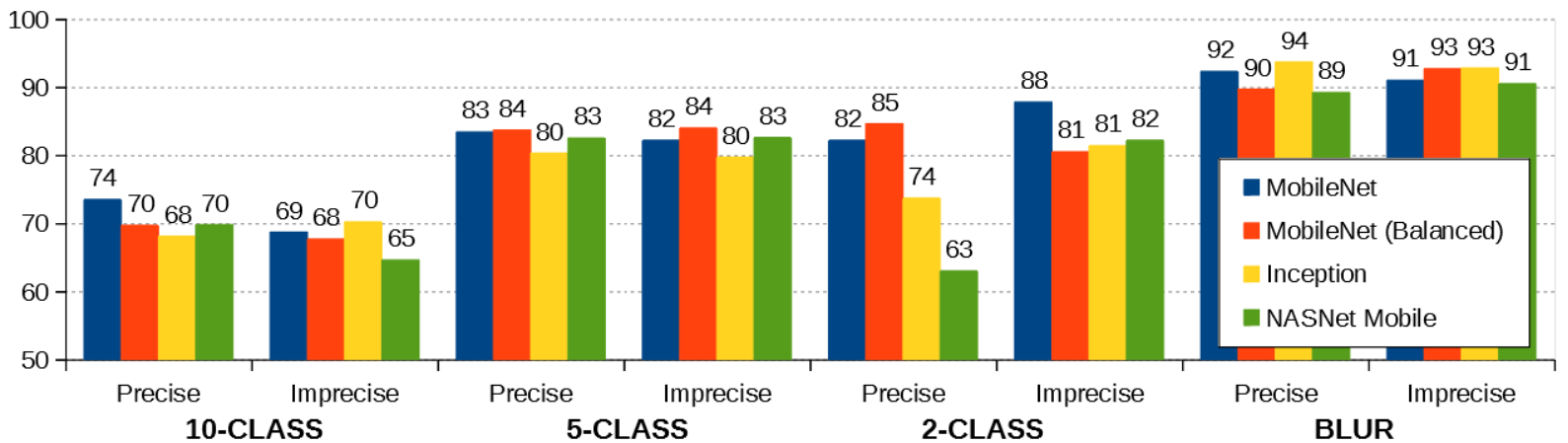}
\caption{F-score (in \%)  for each neural network, for each problem and data set used.}
\label{fig1}
\end{figure}

\begin{figure}[ht]
\centering
\includegraphics[width=0.9\textwidth]{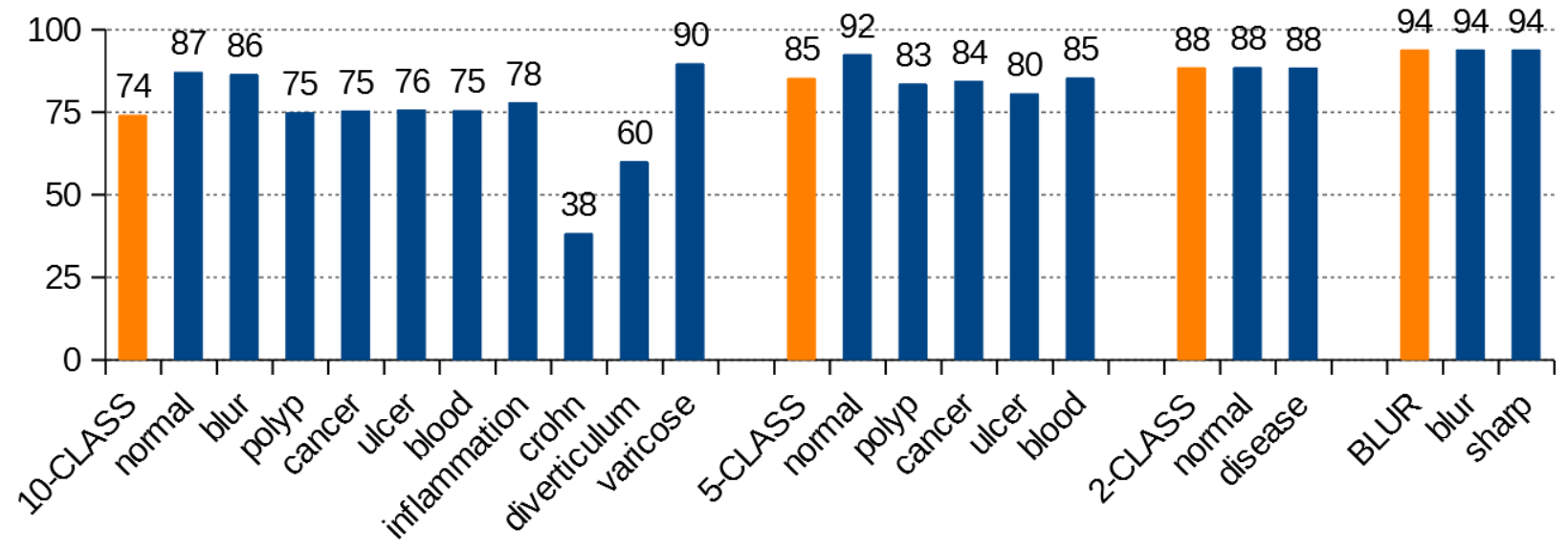}
\caption{F-scores (in \%) for each class in every problem considered, using best neural networks found (orange bar presents the average F-score over all classes).}
\label{fig2}
\end{figure}

As a neural net efficiency metric, F-score based on sensitivity and specificity for each class was used. Final performances for whole network were calculated with macro-averaging (average over F-scores of individual classes). Results on test set for each neural network are presented in Fig. \ref{fig1}. For each problem, the best neural network found was used to calculate F-scores for individual classes in that problem, presented in Fig. \ref{fig2}.

The investigated problems, as predicted, showed different levels of difficulty. The most difficult problem was a 10-class problem, where the recognition of normal tissue, blurry image and varicose reached an F-score of 86--90\%, and for diseases about 75--78\% (except for Crohn's disease and diverticulosis, probably due to a small number of training images).  In the 5-class problem, F-score for diseases reaches 80--85\%, and for healthy tissue 92\%. A better result than in 10-class problem is due to fewer diseases analysed and no classes with a low number of training images. In 2-class problem 88\% effectiveness was achieved in recognizing abnormal and healthy tissue. The lower effectiveness in recognizing healthy images than in 10-class and 5-class problems is due to the fact that in this problem, absolutely all diseases from the data set are analyzed, not just a subset of them. The simplest problem turned out to be the BLUR problem, where an F-score of 94\% was obtained.

The conducted experiments do not indicate major differences in effectiveness between the tested neural networks. On average, MobileNet achieved slightly better results than other networks in most cases, and NASNet-Mobile was the worst. Noteworthy is a two-class problem, in which MobileNet achieved a significantly higher F-score than other networks (88\%). At the same time, it was the only problem where the use of imprecise annotations resulted in significant improvement. The application of additional balancing of counts of images per patient also did not bring significant differences in network operation in any problem.

\onecolumn

\twocolumn

\begin{figure}[ht]
\centering
\begin{subfigure}{.5\textwidth}
  \centering
  \includegraphics[width=.99\linewidth]{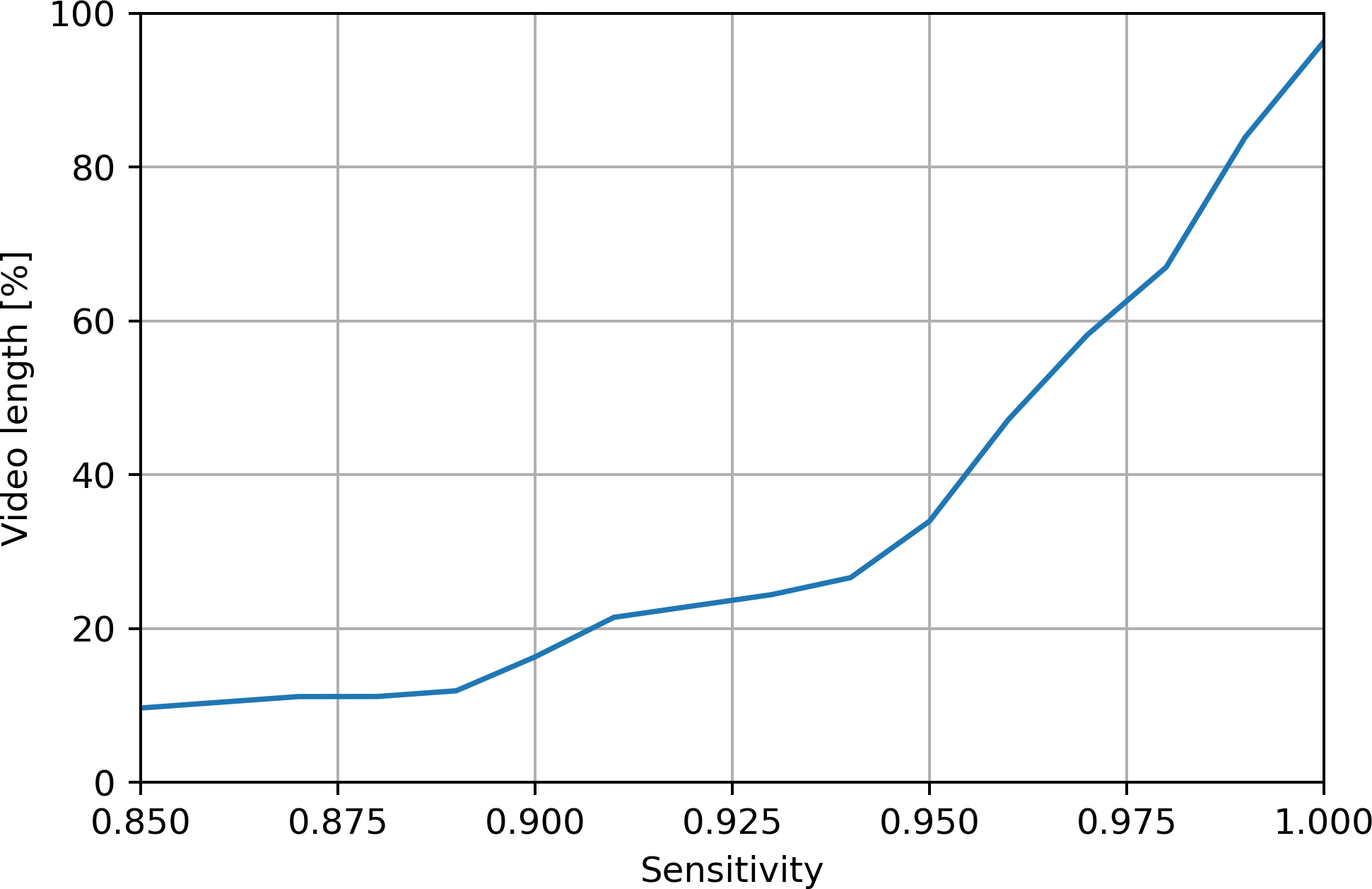}
  \label{fig:sub1}
\end{subfigure}%
\begin{subfigure}{.5\textwidth}
  \centering
  \includegraphics[width=.99\linewidth]{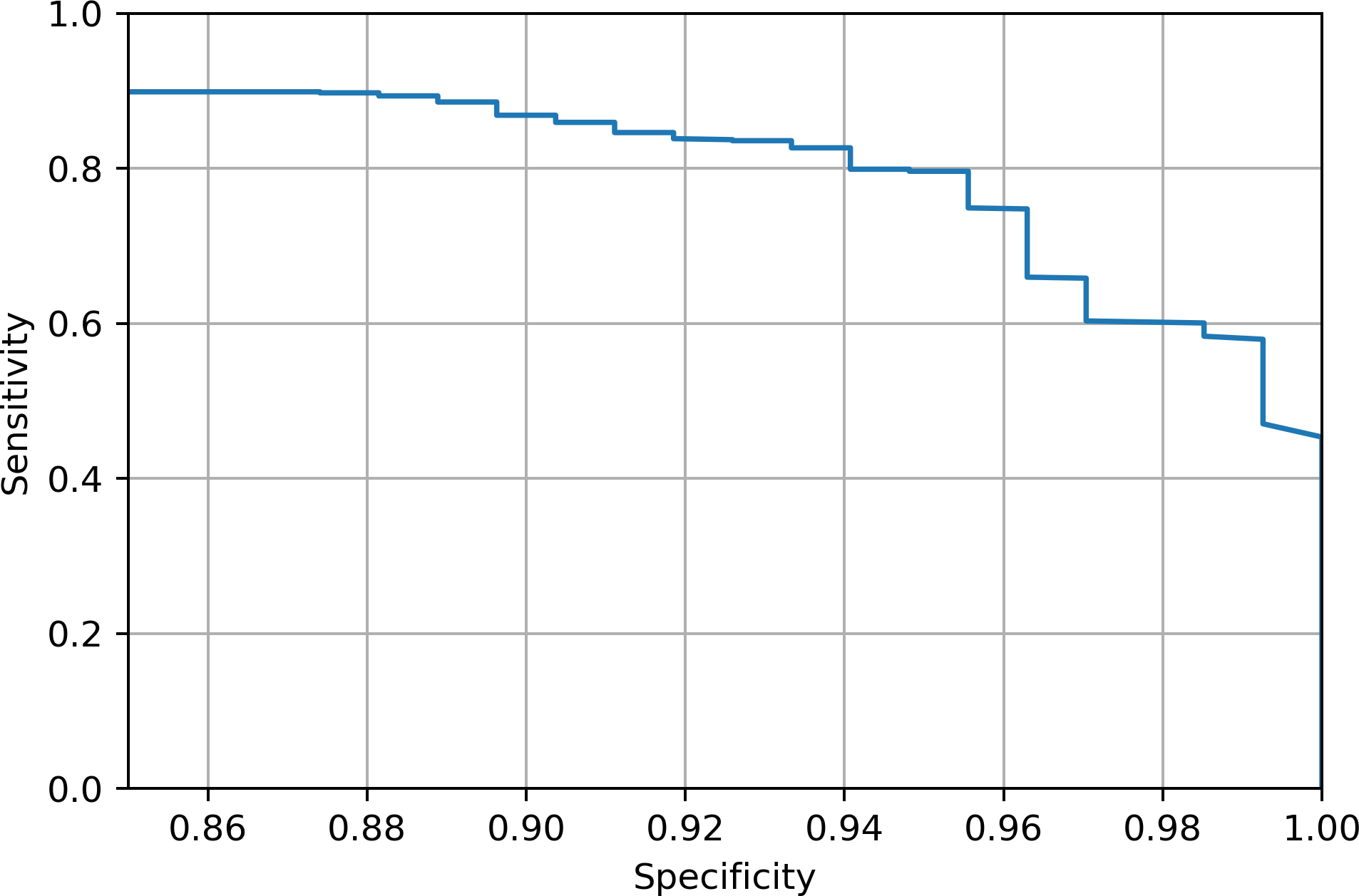}
  \label{fig:sub2}
\end{subfigure}
\caption{Video length after removing healthy images for specified anomaly detection sensitivity (left); sensitivity of anomaly detection for specified specificity (right).}
\label{fig3}
\end{figure}

In addition, we carried out the tests to estimate impact of using the MobileNet network (trained for 2-class problem described before) on video analysis time shortening by physicians. The classifier detects which video frames do not contain any anomalies and removes them, shortening the video length. The operation of the algorithm depends on the parameter of the desired sensitivity, which determines how many potential undetected anomalies the user is able to accept. For a parameter equal to 99\% (which means the requirement to detect 99\% anomalies), the video length has been reduced by 16\%. With the reduction of this requirement, the film was additionally much shorter (by over 80\%). It allows to shorten an examination time from about hours up to a few minutes (in case WCE examination). The relationship between the specified algorithm specificity and the obtained anomaly detection efficiency was also examined. The detailed results are presented in Fig. \ref{fig3}.

\onecolumn

\twocolumn

\section{Conclusions}
In this paper we present a new large data set which could be useful in a training classifiers for computer-based diagnosis of GI tract. We also show a validation study comparing the performance of different CNN models for various class of problems.
The review of the available data sets shows the need to create a uniform, standardized research repository. The availability of large public collections, in particular in the field of gastroscopy, is very limited, and as important as colonoscopy. We believe that the presented ERS data set can be the foundation for such a benchmark platform.
The study shows that access to large available annotated data is needed for a comprehensive validation of GI indications classification, detection and segmentation method and that this might lead to a boost in performance of end-to-end learning methods.
The analysis of the results obtained for each class shows at a quite good performance in multi-class classification and superior at small number of classes, in both full-video sets and still images. 
For the multi-class classification, it shows how important the amount of collected data is (Crohn's disease has only 84 precise images whereas the set of diverticulum contains 135 and the set of polyps - 1192 images and their F-scores are 38, 60 and 86 respectively). 
The use of combination of different methods and algorithms shows that they are close to clinical use because it is able to select and order the most significant images in videos in a short time.


\section*{Acknowledgements}

\ifanon\else
This research was partially funded by grants from National Centre for Research and Development (PBS2/A3/17/2013, Internet platform for data integration and collaboration of medical research teams for the stroke treatment centers).
\fi


\bibliography{ers-2022}


\ifappendix

\onecolumn

\section*{Appendixes}

\begin{figure*}[ht]
    \centering
    \includegraphics[width=\textwidth,height=14cm,keepaspectratio]{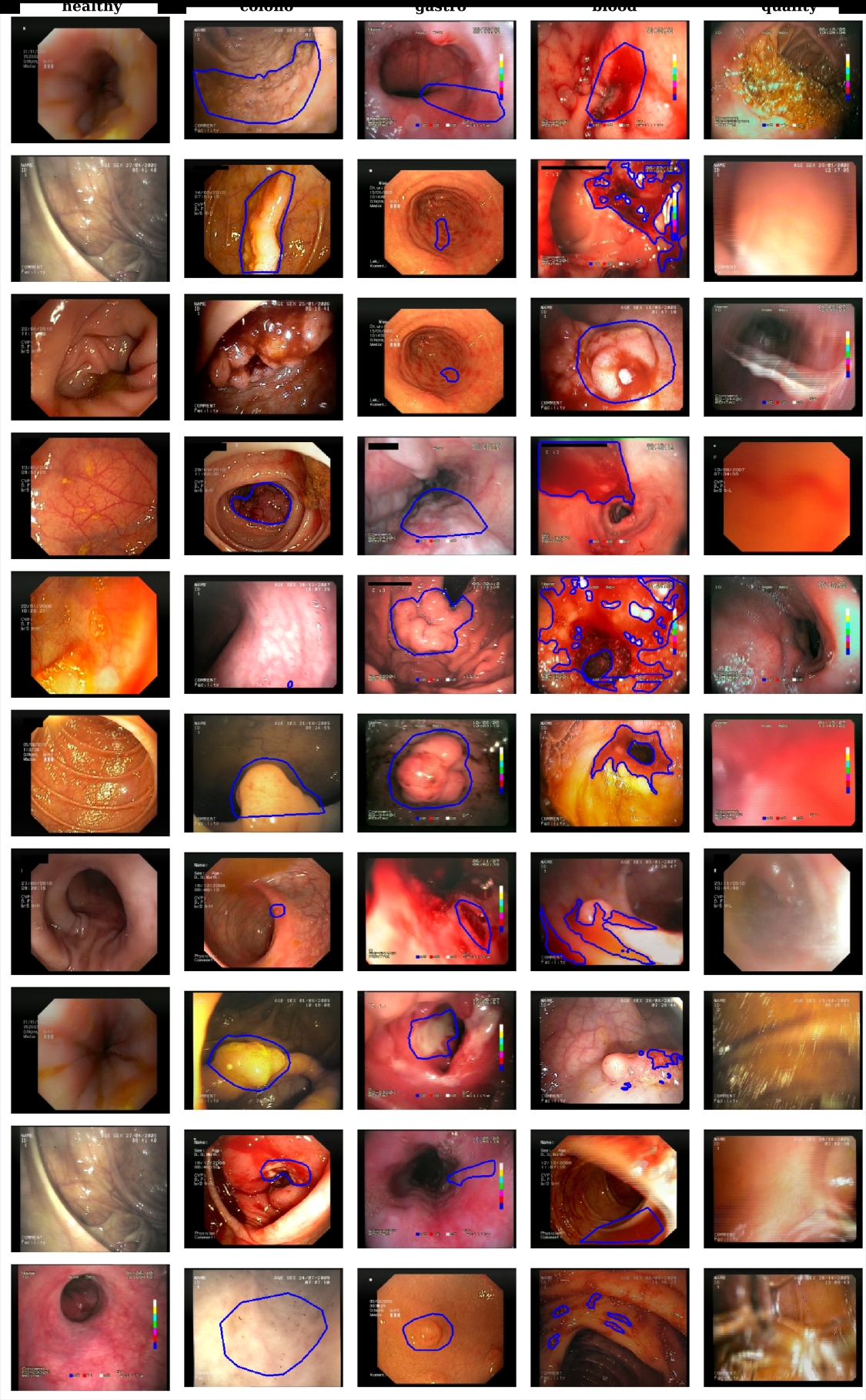}
    \caption{Examples of images (and annotation masks -- as blue contours) in the data set}
    \label{fig:example_images}
\end{figure*}



\tabletextsize

\begin{longtable}{p{1.5cm}|p{0.9cm}|p{0.9cm}|p{0.7cm}p{0.7cm}p{0.7cm}p{0.7cm}p{1.1cm}p{0.7cm}p{0.75cm}p{0.7cm}}
\caption{Problems, class names, and term IDs from the data set used.} \\

&normal&blur&polyp&cancer&ulcer&blood&inflam-mation& Crohn &diverti-culum&varicose\\
\cline{2-11}
\multirow{2}{*}{10-CLASS}&b02, h&
q02, q05, q07, q09&
c22, g14, g31, g45&
c05, g11, g26, g39&
c26, c27, c32, g18, g19, g50, g51, g52&
b01, c02, c05, g05, g08, g19, g39, g51&
c03, c16, c25, c32, c34, g03, g18, g20, g22, g23, g54, g55, g56, g57, g59, g63, g67&
c17, c31, g09, g24, g33, g35, g44, g70&
c12, c13, g28, g29, g40, g41&
g36, g53\\
\hline
&
normal&
polyp&
cancer&
ulcer&
blood&&&&& \\
\cline{2-6}
\multirow{2}{*}{5-CLASS}&b02, h&
c22, g14, g31, g45&
c05, g11, g26, g39&
c26, c27, c32, g18, g19, g50, g51, g52&
b01, c02, c05, g05, g08, g19, g39, g51&&&&& \\
\cline{1-6}

&normal
&disease& & & & & & & & \\
\cline{2-3}
\multirow{2}{*}{2-CLASS}&b02, h&
b01, c, g& & & & & & & & \\
\cline{1-3}
\multirow{2}{*}{BLUR}&
blur&
sharp& & & & & & & & \\
\cline{2-3}
&
q02, q05, q07, q09&
b, c, g, q01& & & & & & & & \\
\cline{1-3}
\end{longtable}



\begin{longtable}{p{1.5cm}|p{0.9cm}|p{0.9cm}|p{0.7cm}p{0.7cm}p{0.7cm}p{0.7cm}p{1.2cm}p{0.7cm}p{0.75cm}p{0.7cm}}
\caption{Number of used images (patients in parentheses) for each problem and variant. }\\

10-CLASS& normal&blur&polyp&cancer&ulcer&blood&inflam-mation& Crohn &diverti-culum&varicose\\
\hline
Precise
&1022 (69)
&589 (75)
&1192 (339)
&688 (151)
&567 (197)
&769 (194)
&805 (267)
&84 (27)
&135 (50)
&252 (85)\\
\hline
Imprecise
&17888 (113)
&28145 (190)
&6648 (339)
&19569 (152)
&3324 (197)
&19717 (195)
&13768 (268)
&372 (27)
&626 (50)
&1111 (85)
\\
\hline
5-CLASS&
normal&
polyp&
cancer&
ulcer&
blood&&&&& \\
\cline{1-6}
Precise
&1022 (69)
&1192 (339)
&688 (151)
&567 (197)
&769 (194)\\
\cline{1-6}
Imprecise
&17888 (113)
&6648 (339)
&19569 (152)
&3324 (197)
&19717 (195)\\
\cline{1-6}

2-CLASS&normal
&disease& & & & & & & & \\
\cline{1-3}
Precise
&1022 (69)
&3980 (1070)\\
\cline{1-3}
Imprecise
&17888 (113)
&46156 (1071)\\
\cline{1-3}
BLUR&
blur&
sharp& & & & & & & & \\
\cline{1-3}
Precise
&589 (75)
&4866 (1091)\\
\cline{1-3}
Imprecise
&28145 (190)
&63869 (1102)\\
\cline{1-3}
\end{longtable}

\normalsize

\twocolumn

\fi 

\end{document}